\documentclass[10pt,twocolumn,letterpaper]{article}

\usepackage{iccv}
\usepackage[accsupp]{axessibility}  %
\usepackage{times}
\usepackage{epsfig}
\usepackage{graphicx}
\usepackage{amsmath}
\usepackage{amssymb}
\usepackage[accsupp]{axessibility} 
\usepackage{multirow}

\usepackage{pifont}
\usepackage[dvipsnames]{xcolor}
\definecolor{limegreen}{HTML}{32CD32}
\newcommand{\cmark}{\textcolor{limegreen}{\ding{51}}}%
\newcommand{\xmark}{\textcolor{red}{\ding{55}}}%

\usepackage{booktabs}
\usepackage{graphbox} %
\usepackage{makecell}
\usepackage{pifont}%

\usepackage{gensymb}
\usepackage{capt-of} %
\usepackage{mathtools} %
\usepackage{multirow}
\usepackage{placeins} %

\usepackage{enumitem}
\setitemize[0]{leftmargin=15pt}

\newenvironment{tight_itemize}{
\begin{itemize}[leftmargin=15pt]
  \setlength{\topsep}{0pt}
  \setlength{\itemsep}{0pt}
  \setlength{\parskip}{0pt}
  \setlength{\parsep}{0pt}
}{\end{itemize}}

\usepackage[pagebackref=true,breaklinks=true,letterpaper=true,colorlinks,bookmarks=false]{hyperref}

\iccvfinalcopy 

\ificcvfinal\fi

\begin{document}

\title{NEMTO: Neural Environment Matting for \\ Novel View and Relighting Synthesis of Transparent Objects}

\author{Dongqing Wang\qquad  Tong Zhang\qquad  Sabine Süsstrunk\\
School of Computer and Communication Sciences, EPFL\\
{\tt\small \{dongqing.wang, tong.zhang, sabine.susstrunk\}@epfl.ch}
\vspace{-2mm}
}

\ificcvfinal\thispagestyle{empty}\fi

\twocolumn[{%
\vspace*{-0.5cm}
\maketitle
}]

\begin{abstract}
We propose NEMTO, the first end-to-end neural rendering pipeline to model 3D transparent objects with complex geometry and unknown indices of refraction. Commonly used appearance modeling such as the Disney BSDF model cannot accurately address this challenging problem due to the complex light paths bending through refractions and the strong dependency of surface appearance on illumination. With 2D images of the transparent object as input, our method is capable of high-quality novel view and relighting synthesis. We leverage implicit Signed Distance Functions (SDF) to model the object geometry and propose a refraction-aware ray bending network to model the effects of light refraction within the object. Our ray bending network is more tolerant to geometric inaccuracies than traditional physically-based methods for rendering transparent objects. We provide extensive evaluations on both synthetic and real-world datasets to demonstrate our high-quality synthesis and the applicability of our method. 
\end{abstract}

\section{Introduction}
\label{sec:intro}

 Modeling transparent objects is important for VR/AR applications as the former are abundant in the real world. Unlike opaque objects with close-to-zero light transmission, transparent objects allow light to pass through. Such refraction and reflection create complex light paths and give transparent objects highly environment-dependent appearances. Consequently, the appearance and geometry of transparent objects are much more entangled than those of opaque objects~\cite{pharr2016physically}. A slight error in object geometry can lead to a global change in appearance~\cite{xu2022hybrid}, as the light path for each ray may thus vary substantially. For these reasons, deriving material and object geometry from images of a transparent object is a highly ill-posed and challenging problem.

\begin{figure}[t]
\begin{center}
   \includegraphics[width=1.1\linewidth]{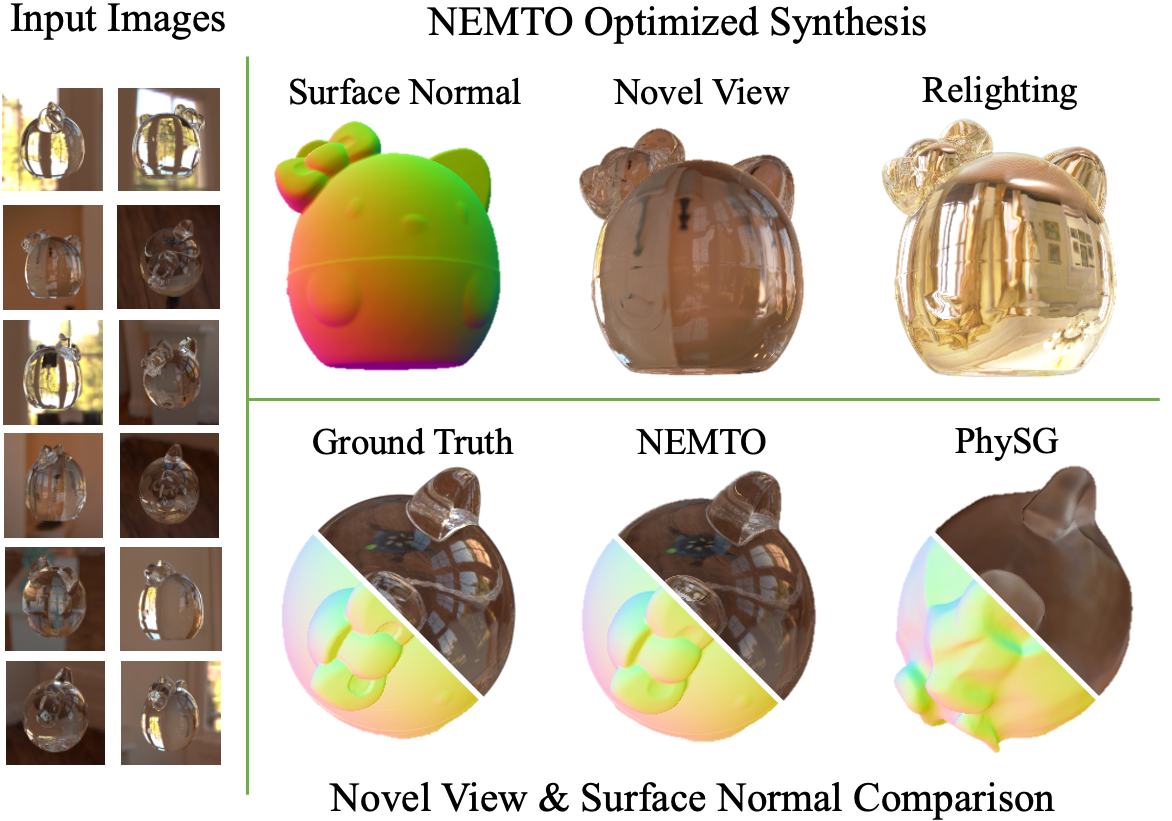}
\end{center}
   \caption{Given as input multi-view images captured under natural illumination, NEMTO is capable of high-quality novel view synthesis and relighting through optimizing an end-to-end neural representation for a transparent object. NEMTO disentangles geometry and illumination-dependent appearance, which previous neural rendering methods, such as PhySG~\cite{zhang2021physg}, cannot. }
\label{fig:teaser}
\end{figure}

Existing work for modeling transparent objects can be classified into two categories. One assumes known indices of refraction (IOR) and reconstructs the complex geometry of transparent objects through either physical devices and structured backlights ~\cite{chuang2000environment, lyu2020differentiable, wetzstein2011refractive, wexler2002image, wu2018full, xu2022hybrid} or neural networks that model geometry with analytical refraction~\cite{li2020through}. The other~\cite{bemana2022eikonal} focuses on optimizing the refractive ray path in the scene without modeling the object surface geometry. However, neither approach is optimal for synthesizing novel views and relighting for transparent objects  with \textit{complex geometry}. In this work, we propose a new framework that combines recent advances in Neural Inverse Rendering~\cite{boss2021nerd, boss2021neuralpil, munkberg2022extracting, zhang2021physg, zhang2021nerfactor, zhang2022invrender} to overcome these limitations.

Traditionally, physically-based rendering follows Snell's Law to render transparent objects. However, object appearance highly depends on geometry estimation, and jointly optimizing both is highly ill-posed. Therefore, our key contribution is incorporating a physically-guided \textit{Ray Bending Network} (RBN) to disentangle object geometry and light refraction. RBN takes the learned geometry~\cite{yariv2020multiview} as prior, and models light refraction by mapping the incoming ray direction directly to the refracted ray direction exiting the object. Our method does not assume a homogeneous refractive index or a fixed number of bounces~\cite{bemana2022eikonal, pharr2016physically}, and models the object's surface with the zero-level set of the Signed Distance Function (SDF). NEMTO thus has the potential to handle a wider range of complex geometry and better adapt to various refractive media than existing transparent object modeling~\cite{bemana2022eikonal, li2020through}. Furthermore, our RBN can improve the estimated geometry by better disentangling it from the appearance of the object than other neural rendering methods~\cite{yariv2020multiview, zhang2021physg}. NEMTO thus makes it practical to model transparent objects in various scenarios, by working with unknown refractive indices and natural environment illumination. 
 Tab.~\ref{tab:baseline} lists the pros and cons of image-based models on novel view and relight synthesis, along with methods focusing on geometry estimation for transparent objects. We identify the first group as our baseline because the second cannot synthesize views without knowing the object IOR. Experiments show that NEMTO can synthesize higher quality novel views and relighting through our representation of transparent objects than all of our baseline methods.

To summarize, our contributions are as follows: 
\begin{tight_itemize}
    \item We propose NEMTO, the first end-to-end method for novel view synthesis and scene relighting for \textit{transparent objects}, shown in Fig.~\ref{fig:teaser}. Our method can disentangle transparent object geometry and appearance. 
    \item We design a physically-guided Ray Bending Network (RBN) for predicting ray paths traversing through the transparent object. The network prediction has better error tolerance for the estimated geometry than analytically calculated refraction.
    \item NEMTO can easily be adapted to real-world transparent objects and achieve high-quality image-based synthesis. 
\end{tight_itemize}
\vspace{-0.4cm}

\section{Related Work}
\label{sec:related}
 \newcommand\RotText[1]{\rotatebox{90}{\parbox{2cm}{\centering#1}}}

\begin{table}[t]
\resizebox{\linewidth}{!}{%
\begin{tabular}{c|cccccccl}
\begin{tabular}[c]{@{}c@{}}\textbf{Methods}\end{tabular} & 
\begin{tabular}[c]{@{}c@{}}{\textbf{A}}\end{tabular} &
\begin{tabular}[c]{@{}c@{}}{\textbf{B}}\end{tabular} 
&\begin{tabular}[c]{@{}c@{}}{\textbf{C}}\end{tabular} & \begin{tabular}[c]{@{}c@{}}{\textbf{D}}\end{tabular} & \begin{tabular}[c]{@{}c@{}}{\textbf{E}}\end{tabular} & \begin{tabular}[c]{@{}c@{}}{\textbf{F}}\end{tabular}&\begin{tabular}[c]{@{}c@{}}{\textbf{G}}\end{tabular} & \textbf{Task} \\
\toprule
NeRF~\cite{mildenhall2020nerf}  & \xmark  & \cmark & \xmark  & \xmark & \cmark & \cmark   & \xmark &
\multirow{6}{*}{\rotatebox{90}{\parbox{2cm}{\centering \textbf{\hspace{3pt} Img-Based Synthesis }}}} \\

Eikonal~\cite{bemana2022eikonal}   & \cmark & \cmark & \xmark & \xmark & \cmark  & \cmark  & \xmark  &  \\

IDR~\cite{yariv2020multiview}  & \xmark  & \cmark   & \xmark   & \cmark   & \cmark    & \cmark  & \xmark  &  \\
 \small{PhySG, ...}~\cite{zhang2021physg, zhang2022invrender}   & \xmark  & \cmark  & \cmark   & \cmark   & \cmark  & \xmark  & \cmark  &  \\

\cmidrule(lr){1-8}
\textbf{NEMTO (Ours)}   & \cmark    & \cmark   & \cmark  & \cmark  & \cmark      & \cmark  & \xmark &  
\multirow{5}{*}{\rotatebox{90}{\parbox{1.2cm}{\centering \textbf{\hspace{0pt} Geo. \\ Est.}}}} \\

\midrule
\cite{wu2018full, lyu2020differentiable, xu2022hybrid} & \cmark & \xmark & \xmark  & \cmark  & \xmark     & \xmark  & \xmark  &  \\

TLG~\cite{li2020through}  & \cmark  & \xmark & \xmark  & \cmark  & \cmark & \xmark   & \xmark  &\\
\bottomrule
\end{tabular}
}
\vspace{-3pt}
\caption{\textbf{Comparison of relevant methods. } The first group focuses on image-based novel view synthesis and relighting, while the second estimates transparent object geometry. \textbf{(A)} can model light refraction for non-opaque objects,  \textbf{(B)} allows direct novel view synthesis \textit{w/ unknown IOR},  \textbf{(C)} allows direct scene relighting \textit{w/ unknown IOR}, \textbf{(D)} can model object surface,  \textbf{(E)} does not require complex setup for image capture, i.e no patterned backlight, turntables, etc.,  \textbf{(F)} can model transparent materials \textit{w/ unknown IOR}, \textbf{(G)} allows estimation of illumination during training. \vspace{-10pt}} 
\label{tab:baseline}
\end{table}

\medskip
\noindent\textbf{Neural Rendering.}
Neural rendering algorithms with implicit scene representation fall into two categories, volume-based and surface-based methods. Volume-based methods, e.g.\ NeRF~\cite{mildenhall2020nerf}, enable photo-realistic novel view synthesis by representing the scene as a Multilayer Perceptron (MLP) based radiance field ~\cite{boss2021nerd, verbin2022refnerf, yariv2021volume}. These methods often cannot distill radiance near the object surface, which is disadvantageous in our case as light refraction strongly relies on ray-surface intersections as prior. Surfaced-based methods~\cite{yariv2020multiview} directly optimize the underlying geometry with SDFs
and estimate object surface with higher accuracy. 

Both representations have evolved to model appearance via the rendering equation \cite{kajiya1986rendering} i.e.,\ to jointly estimate the scene geometry, appearance, and illumination of the scene using existing 2D images ~\cite{bi2020neural, boss2021nerd, verbin2022refnerf, zhang2021physg, zhang2021nerfactor, zhang2022invrender}. These methods assume that objects have opaque surfaces and light paths are non-refractive throughout the scene, and model appearance with the Disney BRDF model~\cite{burley2012physically, karis2013real}. They provide insights into solving the highly ill-posed inverse rendering problem, but cannot work for transparent objects. In our model, we design an MLP for ray refraction prediction to allow modeling of light through non-opaque objects.

\begin{figure*}
\begin{center}
\includegraphics[width=\linewidth]{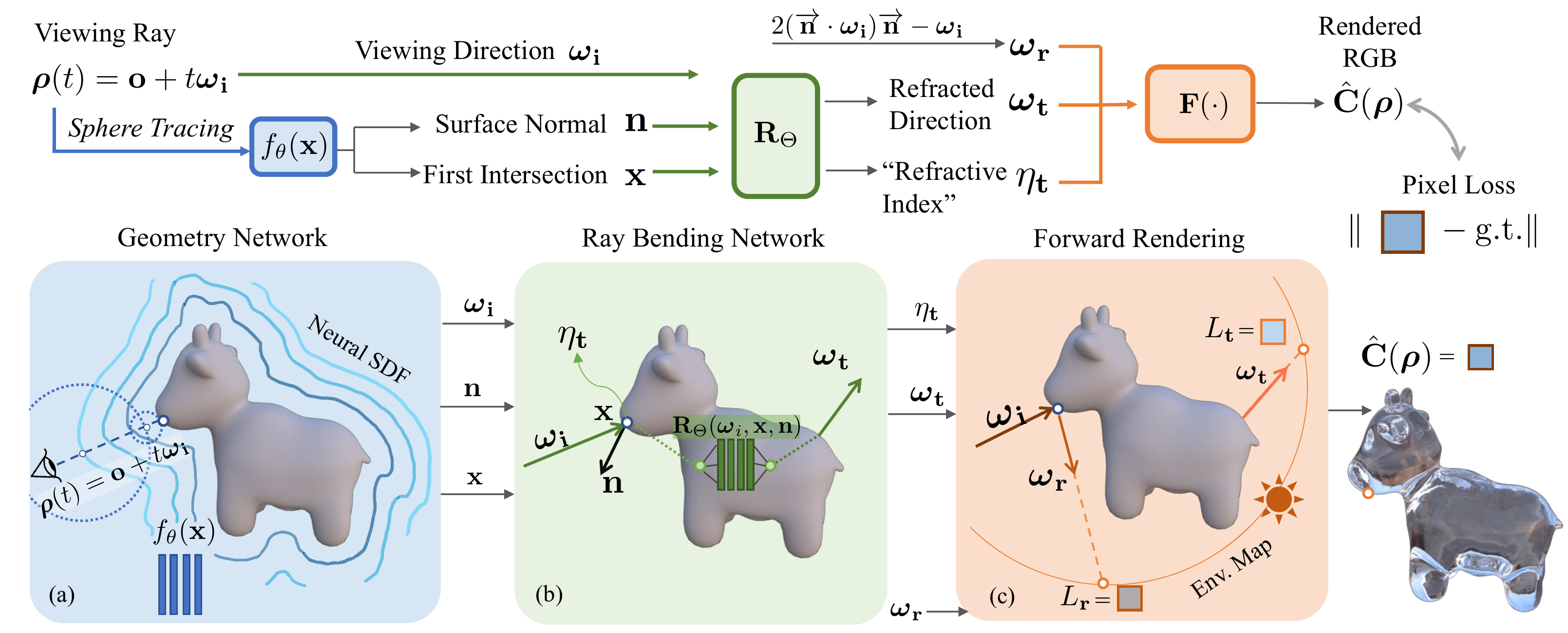}
\end{center}
   \caption{\textbf{Overview of NEMTO framework}. (a) \textit{Geometry Network}. For each viewing ray $\boldsymbol\rho(t) = \mathbf{o} + t \boldsymbol{\omega_\mathbf{i}}$, we query geometry network $f_\theta$ through sphere tracing for the ray-surface intersection. (b) \textit{Ray Bending Network.} We map the viewing direction $\boldsymbol{\omega_i}$ directly to the final refracted ray $\boldsymbol{\omega_\mathbf{t}}$ exiting the object geometry with surface normal $\mathbf{n}$ and intersection $\mathbf{x}$ as prior. As we use an environment map as illumination, the radiance evaluated through refraction only depends on the ray direction, not the location that the ray exits from. (c) \textit{Forward Rendering.} To render $\boldsymbol\rho(t)$,  we analytically calculate reflection direction $\boldsymbol{\omega_\mathbf{r}}$ through $\boldsymbol{\omega_i}$ and $\mathbf{n}$. We then use our physically-inspired rendering algorithm with predicted ``refractive index'' $\eta_\mathbf{t}$ and evaluate the environment map through $\boldsymbol{\omega_\mathbf{t}}$ and $\boldsymbol{\omega_\mathbf{r}}$. }
\label{fig:modelarch}
\end{figure*}

\medskip
\noindent\textbf{Environment Matting.}
Environment matting captures the reflection and refraction of environment light by transparent objects. It represents illumination as texture maps and recovers refraction through pixel-texel correspondence. With a structured backlight as background, the light path through the object in the front can be approximately computed \cite{chuang2000environment, zongker1999environment}. Inspired by traditional environment matting techniques for transparent object shape reconstruction,  Chen \etal~\cite{chen2018tom} design a deep learning network to estimate the environment matting as a refractive flow field. These above methods require a controlled dark room to capture images without ambient light. Wexler \etal~\cite{wexler2002image} develop an environment matting algorithm that works with natural scene backgrounds, but that method requires complex camera setups and structured background light. In our method, we approximate environment matting through a neural network. We directly map camera rays to refracted rays and optimize by comparing projected pixels on the environment maps to ground truth pixels. Our method does not require a complex physical setup and is more adaptive to inaccurate geometry.

\medskip
\noindent\textbf{Transparent Object Modeling.}
Forward rendering of transparent objects is well-understood given Snell's Law. Inversely rendering transparent objects and reconstructing the geometry from images, however, remains challenging. Kutulakos \etal~\cite{kutulakos2008theory} first prove that a ray path through two-interface refractive media can be recovered theoretically. Given this insight, previous methods estimating transparent geometry use controlled setups for light path acquisition such as light field probes~\cite{wetzstein2011refractive}, polarized imagery \cite{huynh2010shape}, X-ray CT scanner \cite{stets2017scene}, and transmission imaging~\cite{kim2017acquiring}.
Wu \etal~\cite{wu2018full} and Lyu \etal \cite{lyu2020differentiable} use turntables in front of static structured backlights to reconstruct geometry, but our work does not require these special setups. 
Neural networks are later introduced with analytical refraction modeling to approach the problem. Li \etal~\cite{li2020through} assume known environment illumination and IOR to only optimize for geometry with one neural network on all kinds of geometry. 
Bemana \etal~\cite{bemana2022eikonal} synthesize novel views for transparent objects with unknown IOR, but only work with simple geometry such as a sphere. Our work doesn't assume known IORs or restrict the light refraction to two bounces. By directly estimating object-specific light refractions, NEMTO synthesizes photo-realistic novel views and relighting for transparent objects on casually-captured images. 

\vspace{-0.2cm}
\section{Method }
\label{sec:method}

\subsection{Problem Formulation }
Each input dataset contains $N$ multi-view images $\{\mathbf{I}_n\}^N_{n=1}$ and $N$ object masks $\{\mathbf{Q}_n\}^N_{n=1}$ of a transparent object with unknown IOR,  and an environment map $\Gamma:\mathbb{R}^{H\times W\times3}$ for scene illumination. Object masks and the environment map can be generated by preprocessing image data, which we detail in the supplementary. 
We assume the images are captured under static distant illumination represented by the environment map. Separately estimating illumination can compensate for entanglement between geometry and lighting, as transparent object appearance is highly correlated to scene illumination. We relax the assumptions that light bounces twice within the transparent media in Li \etal~\cite{li2020through} and do not require two reference points on each ray~\cite{kutulakos2008theory}. 
The model architecture is summarized in Figure~\ref{fig:modelarch}.

\subsection{Geometry Network }
\label{subsec:geometry}
 We adopt implicit signed distance functions (SDFs)~\cite{yariv2020multiview} to represent object geometry due to their adaptive resolution and memory efficiency~\cite{park2019deepsdf}. Specifically, we represent the geometry as a zero-level set $z_\theta$ of an MLP neural network $f_\theta: \mathbb{R}^3 \rightarrow \mathbb{R}$ mapping a 3D location $\mathbf{x}$ to its SDF value $z \in \mathbb{R}$, $\theta$ being its optimizable weights. Concretely, $z_\theta = \{\mathbf{x} \in \mathbb{R}^3 \; | \;f_\theta(\mathbf{x}) = 0 \}$. We optimize geometry with silhouette loss through $\{\mathbf{Q}_n\}^N_{n=1}$ and regularize the neural SDF with IGR regulariztion~\cite{icml2020_2086} for a smooth and realistic object surface, detailed in Sec.~\ref{subsec:opt}. 
 
 To find the intersection point $\mathbf{x}$ between the viewing ray $\boldsymbol\rho(t) = \mathbf{o} + t \boldsymbol{\omega}_\mathbf{i}$ and implicit geometry surface, we perform ray casting through sphere tracing~\cite{hart1996sphere} starting from the intersection between  $\boldsymbol\rho(t)$ and the bounding box of the object. 
 The surface normal of an SDF at $\mathbf{x}$ is directly given by its first order derivative: $\mathbf{n}=\nabla_{\mathbf{x}}f$, therefore enabling gradient flow between normal and geometry optimization. 

\subsection{Ray bending Network }

To render each viewing ray through the transparent object in a physically-plausible manner, we design our model to be consistent with the rendering equation \cite{kajiya1986rendering} and physically-based material model. 

 \medskip
\noindent\textbf{Light Refraction and Reflection Modeling. }
To model a transparent object, 
 one can analytically calculate the light path of an incident ray through the object given the perfect specular reflection and transmission exhibited by its smooth dielectric materials~\cite{pharr2016physically}. Specifically, all scattering of radiance for a given ray shares a single outgoing direction.  
 
As shown in Fig.~\ref{fig:modelarch}. (b), for a 3D point $\mathbf{x}$ on object surface, we denote the incoming ray direction as $\boldsymbol{\omega}_\mathbf{i}$, the unit surface normal as $\mathbf{n}$, the angle between $\boldsymbol{\omega}_\mathbf{i}$ and $\mathbf{n}$ as $\beta_\mathbf{i}$. Likewise denote reflected and transmitted ray direction as $\boldsymbol{\omega}_\mathbf{r}$ and $\boldsymbol{\omega}_\mathbf{t}$ with $\beta_\mathbf{r}$ and $\beta_\mathbf{t}$. Analytically, $\beta_\mathbf{r} = \beta_\mathbf{i}$, and according to Snell's Law, 
\begin{equation}\label{eqn:transmitangle}
\eta_\mathbf{t} \sin{\beta_\mathbf{t}} = \eta_\mathbf{i} \sin{\beta_\mathbf{i}},
 \end{equation}
 where $\eta_\mathbf{i}$ and $\eta_\mathbf{t}$ are defined as the indices of refraction of the medium through which the incoming and outgoing ray travels, respectively. The analytical \textit{reflected} ray direction is expressed as: \vspace{-0.3cm}
 \begin{equation}\label{eqn:reflectdir}
\boldsymbol{\omega}_\mathbf{r} = 2(\mathbf{n}\cdot\boldsymbol{\omega}_\mathbf{i})\mathbf{n} - \boldsymbol{\omega}_\mathbf{i}, 
 \end{equation}
 while the analytical \textit{refracted} ray direction is formulated as \vspace{-0.2cm}
\begin{equation}\label{eqn:transmitdir}
\boldsymbol{\omega}_\mathbf{a} = -\frac{\eta_\mathbf{i}}{\eta_\mathbf{t}} (\boldsymbol{\omega}_\mathbf{i} - (\boldsymbol{\omega}_\mathbf{i} \cdot \mathbf{n})\mathbf{n}) - \mathbf{n}\sqrt{1 - (\frac{\eta_\mathbf{i}}{\eta_\mathbf{t}})^2 (1 - (\boldsymbol{\omega}_\mathbf{i} \cdot \mathbf{n})^2 )}.
 \end{equation}

\vspace{-0.2cm}
\medskip
\noindent\textbf{Neural Environment Matting.}
The accuracy of the analytically evaluated refracted light direction is highly correlated to the quality of the estimated geometry, which makes it difficult to simultaneously optimize the geometry and light refraction through the object. 
To overcome this issue, we discard the analytical solution $\boldsymbol{\omega}_\mathbf{a}$ for refraction modeling in Eq.~\eqref{eqn:transmitdir} and utilize a neural environment matting method. Our ray bending network (RBN) directly estimates $\boldsymbol{\omega}_\mathbf{t}$, mapping incident rays intersecting the scene object to the final refracted ray direction, thereby learning how the transparent object refracts environment light and implicitly represents the refractive index through the network.  

As shown in Fig.~\ref{fig:modelarch} (b), our modeling of light refraction passing through transparent media is expressed as a function $\mathbf{R}$, which takes as input the viewing direction $\boldsymbol{\omega}_\mathbf{i}$, the first intersection point $\mathbf{x}$ between the viewing ray $\boldsymbol\rho(t) = \mathbf{o} + t \boldsymbol{\omega}_\mathbf{i}$ and the implicit geometry surface, and the surface normal $\mathbf{n}$ of point $\mathbf{x}$. The function output is the refracted direction $\boldsymbol{\omega_\mathbf{t}}$ for the viewing ray $\boldsymbol\rho$ exiting the geometry and the ``refractive index'' $\eta_\mathbf{t}$ at $\mathbf{x}$. By incorporating $\mathbf{x}$ and $\mathbf{n}$ as priors, we can account for complex viewing effects, such as total internal reflection, which depends on the angle between the surface normal, viewing direction and the concavity of the geometry. We approximate this function using a Multi-Layer Perceptron (MLP) network $\mathbf{R}_\Theta: (\boldsymbol{\omega}_\mathbf{i}, \mathbf{x}, \mathbf{n}) \rightarrow (\boldsymbol{\omega}_\mathbf{t}, \eta_\mathbf{t})$. To handle high-frequency ray refractions, we apply positional encoding~\cite{tancik2020fourfeat} to the viewing direction and surface intersection.

Our light path modeling stems from two important intuitions inspired by traditional environment matting techniques for transparent objects: (1) we assume the scene to contain a single object that is transparent, and we assume the contribution of radiance on each ray segment exiting the object is negligible except for one main refracted ray; (2) the scene illumination will be modeled with an environment map and each ray comes from an infinite distance. From these observations, the final evaluated radiance of each camera ray that intersects with the scene object only depends on its direction upon exiting the object. 
In the next section, we present our differentiable forward rendering algorithm to work with these assumptions. 

\subsection{Forward Rendering}

The recursive hemispherical integral of the rendering equation for evaluating each viewing ray does not have a closed-form solution and has to be numerically evaluated with the Monte Carlo method~\cite{veach1998robust}. We provide an approximate evaluation of the final radiance as the combination of reflected radiance at incident ray and refracted radiance at outgoing ray through the Fresnel term. Our rendering module is differentiable and designed for physical plausibility.

For transparent objects with smooth surfaces, the view-dependent reflected radiance $L_r$ at each incident point is only dependent on the the reflected ray direction \cite{pharr2016physically, verbin2022refnerf}:\vspace{-0.2cm}
\begin{equation}\label{eqn:reflectrad}
L_r \propto \int f_r(\boldsymbol\omega_\mathbf{r}, \boldsymbol\omega_\mathbf{i}) 
L_i(\boldsymbol{\omega}_\mathbf{i})d\boldsymbol{\omega}_\mathbf{i} = E(\boldsymbol{\omega}_\mathbf{r}),\vspace{-0.3cm}
\end{equation} while $\boldsymbol\omega_\mathbf{i}$ is related to $\boldsymbol{\omega_r}$ through Eqn.~\eqref{eqn:reflectdir}. We propose the assumption that an analogous relationship exists between refracted radiance at the incident location and the final refracted ray direction. Specifically,\vspace{-0.3cm}
\begin{equation}\label{eqn:transmitrad}
L_t \propto \int f_r(\boldsymbol{\omega}_\mathbf{t}, \boldsymbol{\omega}_\mathbf{i}) 
L_i(\boldsymbol{\omega}_\mathbf{i})d\boldsymbol{\omega}_\mathbf{i} = E(\boldsymbol{\omega}_\mathbf{t}),\vspace{-0.3cm}
\end{equation}where $E: \mathbb{R}^3 \rightarrow \mathbb{R}^3$ maps unit ray direction $\boldsymbol\omega_{(\cdot)}$ to a 3-channel RGB color. We reference the RGB value on our estimated environment map by the 2D coordinate obtained through texture mapping from the viewing direction. 

We use the Fresnel equation to compute the incident energy split between reflection and transmission, therefore the reflected and transmitted radiance is proportional to the incident ray radiance at the surface intersection. 
For unpolarized transparent objects, the Fresnel reflectance is given by\vspace{-0.3cm}
\begin{equation}\label{eqn:r}
r_{\parallel} = \frac{\eta_\mathbf{t}\cos{\beta}_\mathbf{i} - \eta_\mathbf{i}\cos{\beta}_\mathbf{t}}{\eta_\mathbf{t}\cos{\beta}_\mathbf{i} + \eta_\mathbf{i}\cos{\beta}_\mathbf{t}},
\end{equation}
\begin{equation}\label{eqn:p}
 r_{\perp} = \frac{\eta_\mathbf{i}\cos{\beta}_\mathbf{i} - \eta_\mathbf{t}\cos{\beta}_\mathbf{t}}{\eta_\mathbf{i}\cos{\beta}_\mathbf{i} + \eta_\mathbf{t}\cos{\beta}_\mathbf{t}},
\end{equation} where $r_{\parallel}$ gives the reflectance for parallel polarized light, and $r_{\perp}$ is the reflectance for perpendicular polarized light. Since we assume light to be unpolarized, the Fresnel reflectance $F_r$ can be analytically written as, \vspace{-0.3cm}
\begin{equation}\label{eqn:fresnel}
F_r = \frac{1}{2}(r_{\parallel}^2 + r_{\perp}^2).
\end{equation}
 Due to the conservation of energy, the energy transmittance $F_t$ is therefore given by $F_t = 1 - F_r$. 

The final radiance for a given ray $\boldsymbol{\rho}$ is then evaluated as: \vspace{-0.3cm}
\begin{equation}\label{eqn:radiance}
\hat{\mathbf{C}}(\boldsymbol\rho) = F_r\odot E(\boldsymbol{\omega}_\mathbf{r}) + \frac{\eta_\mathbf{i}^2}{\eta_\mathbf{t}^2}(1 - F_r) \odot E(\boldsymbol{\omega}_\mathbf{t}),
\end{equation}with $\odot$ denotes element-wise multiplication, and $\eta_\mathbf{i}=1.00028$ being the IOR for air.

\subsection{Optimization}
\label{subsec:opt}
As the joint optimization of geometry and light refraction of transparent objects is highly ill-posed, we enforce different priors to generate visually plausible solutions. 

For each batch, we sample the set of all pixels $P$ from the input image dataset to get $M$ patches, each of $m\times m$ neighboring pixels. Therefore a training batch contains $m^2M$ pixels we denote as $P_M \subset P$. $P_M$ can be subdivided into two non-overlapping sets of pixels $P_{M_i}$ and $P_{M_o}$ depending on whether the pixel contains the object or not, $|P_{M_o}| + |P_{M_i}| = m^2M$. Each $p_{\boldsymbol\rho} \in P_M$ is rendered by one camera ray $\boldsymbol\rho(t) = \mathbf{o} + t \boldsymbol{\omega_\mathbf{i}}$ with origin $\mathbf{o}$ and viewing direction $\boldsymbol{\omega_\mathbf{i}}$. We apply masked rendering and mask out non-intersecting rays for the loss functions. Specifically, through sphere tracing, if $\boldsymbol\rho$ hits the object surface, $p_{\boldsymbol\rho} \in P_{M_i}$, otherwise $p_{\boldsymbol{\rho}} \in P_{M_o}$ and its rendered radiance $\hat{\mathbf{C}}(\boldsymbol\rho) = 0$. 

The Pixel loss for $P_M$ is obtained through the ground truth RGB $\Tilde{\mathbf{C}}_p$ for $p_{\boldsymbol\rho} \in P_{M_i}$ and rendered radiance $\hat{\mathbf{C}}(\boldsymbol\rho)$,
\begin{equation}
\mathcal{L}_{\textrm{pix}} = \frac{1}{|P_{M_i}|}\sum_{p_\mathbf{r}\in P_{M_i}}\lVert \hat{\mathbf{C}}(\boldsymbol\rho)- \Tilde{\mathbf{C}}_p \rVert _1.
\label{eq:img_loss}
\vspace{-0.3cm} 
\end{equation}

For ray refraction estimation, we use two losses: $\mathcal{L}_{\textrm{rg}}$ for each ray $\boldsymbol\rho$ that hits the object with viewing direction $\boldsymbol\omega_\mathbf{i}$ to guide its refracted direction $\boldsymbol\omega_\mathbf{t}$ exiting the object toward the analytical solution $\boldsymbol\omega_\mathbf{a}$ obtained by Eq.~\eqref{eqn:transmitdir} through cosine similarity,
\begin{equation}
\mathcal{L}_{\textrm{rg}} = \frac{1}{|P_{M_i}|}\sum_{\boldsymbol\omega_{\mathbf{i}}: p_{\boldsymbol\rho} \in P_{M_i}}
[ 1 - \frac{\boldsymbol\omega_\mathbf{t} \cdot \boldsymbol\omega_\mathbf{a}}{\max(\lVert\boldsymbol\omega_\mathbf{t}\rVert_2 \cdot \lVert\boldsymbol\omega_\mathbf{a}\rVert_2, 0)} ]. 
\label{eq:rg_loss}
\end{equation} We employ a weight decaying strategy on $\mathcal{L}_{\textrm{rg}} $ to provide initial physically-guided supervision for RBN on refraction. Additionally, we utilize a patch-based loss term $\mathcal{L}_{\textrm{rs}} $ for ray hits to encourage locally smooth refraction directions, as we assume that most transparent objects are locally smooth given their smooth dielectric material. For each $m\times m$ patch containing only ray hits, we penalize the mean of variance on the estimated refraction directions. Along with $\mathcal{L}_{\textrm{pix}}$, the RBN can better compensate the inaccuracy on estimated geometry than strictly following analytical solution. 

For geometry optimization, we employ a silhouette loss from input masks $\{\mathbf{Q}_n\}^N_{n=1}$ and focus on non-intersecting rays. For each ray $\boldsymbol\rho: p_{\boldsymbol\rho} \in P_{M_o}$, we uniformly sample K points on $\boldsymbol\rho$ within the object bounding box and query $f_\theta$ for the minimal SDF value $z_k$ to penalize on ray miss 
with the cross-entropy $\textrm{CE}_\alpha$ loss with logit parameterized by $\alpha$,
\begin{equation}
\mathcal{L}_{\textrm{sil}} = \frac{1}{|P_{M_o}|}\sum_{p_{\boldsymbol\rho}\in P_{M_o}} \textrm{CE}_\alpha(z_k).
\label{eq:mask_loss}
 \vspace{-0.3cm}
\end{equation}
 In order to impose a constraint on the learned function $f_\theta$ to ensure that it is an approximate SDF, we randomly sample $V$ points $\{\mathbf{v_i}\}^V_{i=1}$ inside the bounding box and adopt the loss by Implicit Geometric Regularization (IGR)~\cite{icml2020_2086}:
\begin{equation}
\mathcal{L}_{\textrm{e}} = \mathbb{E}_\mathbf{v}(\lVert \nabla_\mathbf{v}f_\theta \rVert_2 -1)^2 .
\label{eq:eik_loss}
\end{equation}

The overall loss function of our optimization is defined as the weighted sum of each loss with $\lambda_{(\cdot)}$ as weight terms: 
\vspace{-0.3cm}
\begin{equation}
    \begin{split}
    \mathcal{L} &= \lambda_\textrm{pix} \mathcal{L}_\textrm{pix} 
     + \lambda_{\textrm{e}} \mathcal{L}_{\textrm{e}}  + \lambda_\textrm{sil} \mathcal{L}_\textrm{sil}\\
    &+ \lambda_{\textrm{rg}} \mathcal{L}_{\textrm{rg}}  + \lambda_\textrm{rs} \mathcal{L}_\textrm{rs}.
    \label{eq:overall_loss}
    \end{split}
\end{equation}

\section{Experiments}
\label{sec:exp}
\begin{figure*}
\begin{center}
\includegraphics[width=.95\linewidth]{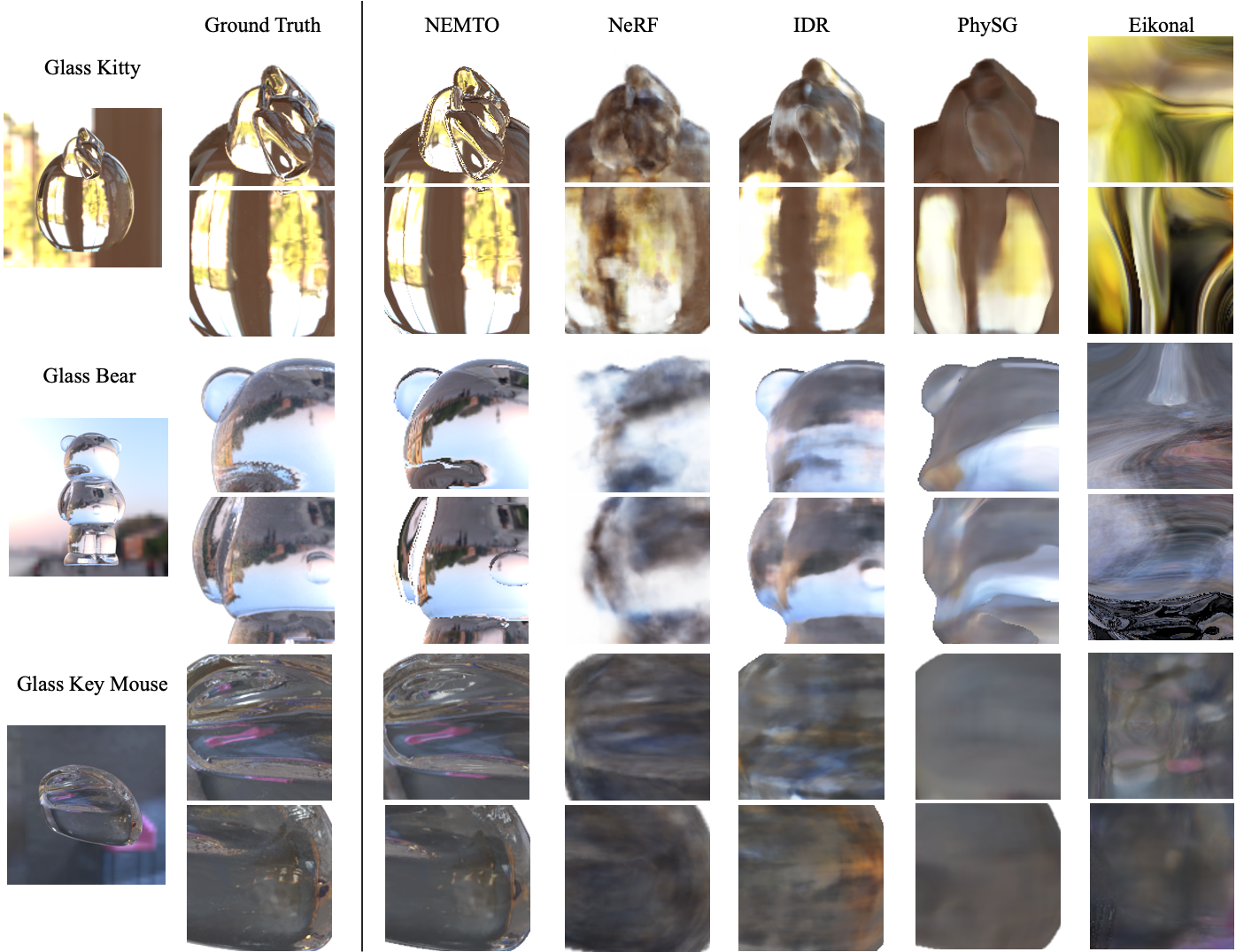}
\end{center}
\vspace{-0.4cm}
   \caption{\textbf{Qualitative comparison with baseline methods on Novel View Synthesis. } We compare our novel view synthesis on transparent objects with the methods that we identify as most relevant to ours, NeRF~\cite{mildenhall2020nerf}, Eikonal Field~\cite{bemana2022eikonal}, IDR~\cite{yariv2020multiview}, and PhySG~\cite{zhang2021physg}. Our method outperforms the others on the high-frequency details caused by ray refraction. }
   \vspace{-0.4cm}
\label{fig:novelviewsyn}
\end{figure*}

\subsection{Synthetic Data Evaluation}
\vspace{-0.2cm}
\medskip
\noindent\textbf{Datasets. }We use the 4 mesh objects of kitty, cow, bear, and key-mouse from~\cite{Xing2022drot, zhang2021physg}, and render each object with the smooth dielectric BSDF model with Mitsuba 3 \cite{jakob2022mitsuba3} under an environmental light emitter. For synthetic dataset evaluations we set interior IOR to 1.4723 for glass and exterior IOR to 1.00028 for air. We also create datasets with interior IOR set to 1.2, and 2.4 for ablation studies. We uniformly sample 200 camera poses on the upper hemisphere around each object following the Fibonacci lattice and randomly assign 100 each for training and testing. We obtain object masks through data pre-processing~\cite{remove_bg}.    

\vspace{-0.2cm}
\medskip
\noindent\textbf{Baseline. }As discussed in Sec.~\ref{sec:related} and shown in Tab.~\ref{tab:baseline}, no other work studies the same problem as ours, i.e., modeling refraction for transparent objects with complex geometry by neural networks for novel view and relighting synthesis. We, therefore, classify our baselines based on different tasks: \textbf{NeRF}~\cite{mildenhall2020nerf} and \textbf{Eikonal Fields}~\cite{bemana2022eikonal} on novel view synthesis; \textbf{IDR}~\cite{yariv2020multiview} on novel view synthesis and geometry reconstruction; \textbf{PhySG}~\cite{zhang2021physg} on novel view and relighting synthesis, and geometry reconstruction.  As geometry is not our aimed task to improve, extracted mesh quality is only to show that RBN effectively disentangles geometry and ray refraction on appearance. We do not include comparisons with volume-based neural relighting methods~\cite{boss2021nerd, verbin2022refnerf} as they share the same appearance model with PhySG.

\begin{figure}
\begin{center}
\includegraphics[width=0.8\linewidth]{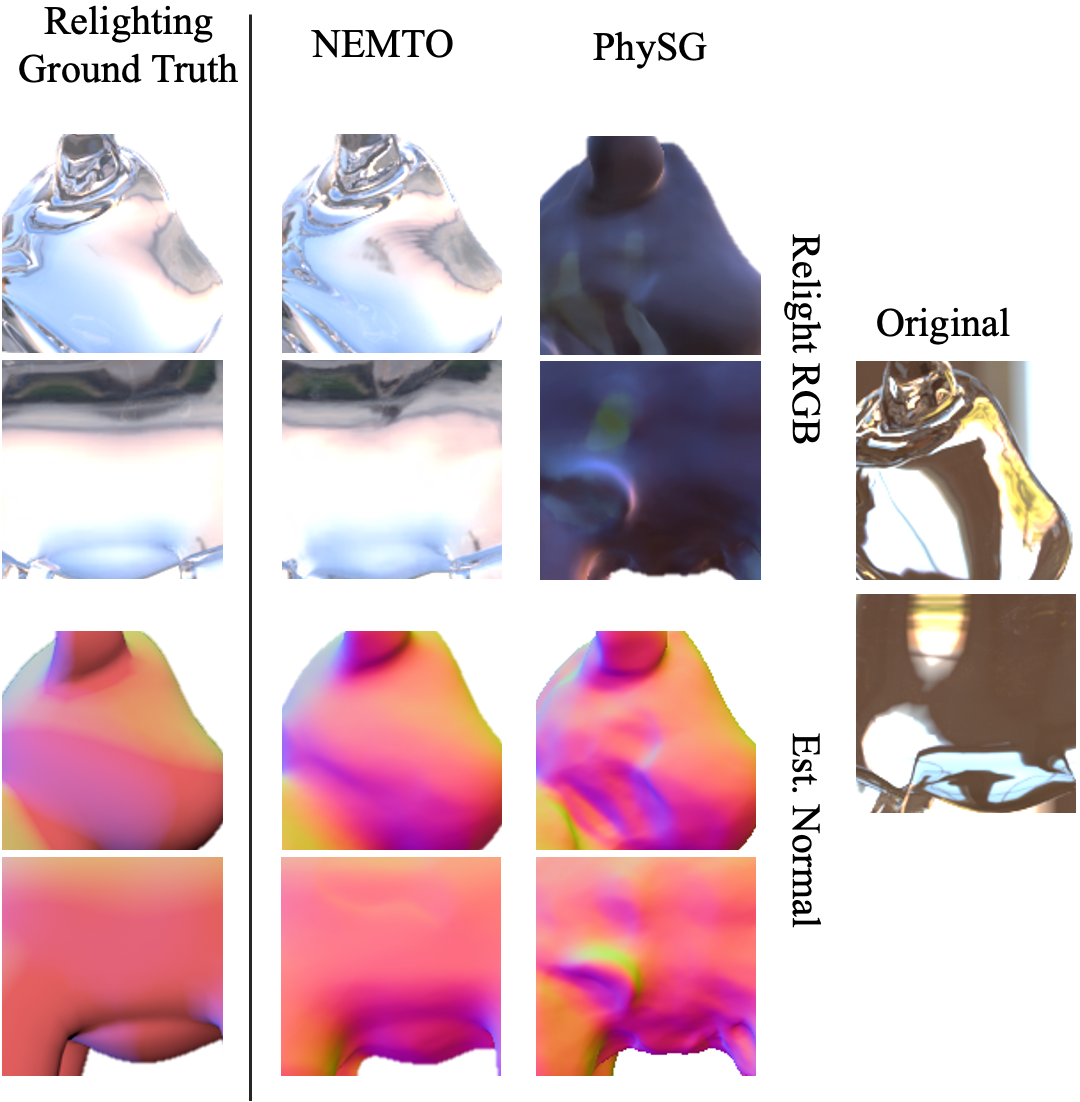}
\end{center}
\vspace{-0.5cm}
   \caption{\textbf{Qualitative results on Relighting for synthetic datasets. } We show that our network can faithfully relight the object with unseen environment illumination, unlike PhySG~\cite{zhang2021physg}. }
\label{fig:relightsyn}
\vspace{-0.4cm}
\end{figure}

\vspace{-0.2cm}
\medskip
\noindent\textbf{Novel View Synthesis.} 
A qualitative comparison of our method and baseline methods is shown in Fig.~\ref{fig:novelviewsyn}. NeRF and Eikonal Fields model object appearance as MLP-based volume and cannot distill radiance properly around the object surface. However, when modeling refractive objects with complex geometry, it is important to locate the surface for accurate refraction direction estimation. Eikonal fields relies on user-defined bounding boxes, resulting in failure cases where the opaque scene and the refractive part cannot be separated. Meanwhile, IDR and PhySG are surface-based methods, but IDR models appearance as a light field~\cite{wood2000surface} and cannot correctly interpolate the high-frequency change of the refracted environment illumination on object appearance. PhySG uses Disney BSDF~\cite{burley2012physically} which does not work for non-opaque objects and therefore fails to correctly disentangle geometry and appearance. 

We report quantitative evaluation on novel view synthesis with metrics including PSNR, SSIM, and LPIPS~\cite{zhang2018perceptual} through testing on held-out images in Tab.~\ref{tab:quanti}. Our method significantly outperforms all of our baselines on synthesizing novel views for accurately modeling the refraction direction of each ray intersected with geometry. 

\medskip
\noindent\textbf{Relight Synthesis.} 
We provide a qualitative comparison of relighting synthesis in Fig.~\ref{fig:relightsyn}. As the environment map used during training is natural and unstructured unlike in prior works~\cite{lyu2020differentiable, wu2018full}, many pixels share similar radiance, but our learned refractions are not overfitted on the training illumination; they are aligned with the true refractions. We relight each scene with an unseen environment map to test the correctness of the object refraction. PhySG fails on this task as it does not model refractive material, resulting in incorrect appearance decomposition~\cite{burley2012physically}.  We report quantitative evaluation w.r.t ground truth relighting in Tab.~\ref{tab:quanti}.

\begin{table}
  \centering
  \scalebox{0.9}{
  \begin{tabular}{@{}lcccc@{}}
    \toprule
    Synthetic & \multicolumn{4}{c}{$\downarrow$Chamfer $L_1(10^{-3})$} \\
    \midrule
    Method & Kitty & Bear & Key Mouse & Cow\\
    \cmidrule{2-5}
    IDR \cite{yariv2020multiview}& 4.30 & 3.66 & 3.70 & 11.66 \\
    PhySG \cite{zhang2021physg} & 87.67 & 67.43 & 31.61 & 52.17 \\
    \textbf{NEMTO} & \textbf{2.22} & \textbf{1.71} & \textbf{2.27} & \textbf{2.60} \\
    \bottomrule
    \vspace{-0.2cm}
  \end{tabular}
    }
  \caption{\textbf{Quantitative evaluation on recovered meshes of synthetic datasets.} We report the chamfer distance metric~\cite{cdcode} on g.t. mesh versus extracted meshes as a quantitative measure for reconstructed geometry quality. NEMTO achieves better results than baseline methods that models object surfaces. }
  \label{tab:chamfersyn}
  \vspace{-0.4cm}
\end{table}

\vspace{-0.2cm}
\medskip
\noindent\textbf{Disentanglement on Geometry and Appearance.}
We evaluate our extracted geometry on synthetic datasets with ground truth mesh through the Chamfer distance metric and compare our geometry with those of surfaced-based methods. In Fig~\ref{fig:relightsyn}, the geometry of PhySG is entangled with surface appearance, i.e. the appearance under the original illumination is imprinted on the surface and raised geometry. Tab.~\ref{tab:chamfersyn} shows that IDR does better than PhySG, though still worse than ours. Our geometry and refracted appearance are better separated due to our modeling of ray refraction and optimizations. 

\begin{figure}[t]
\begin{center}
   \includegraphics[width=0.85\linewidth]{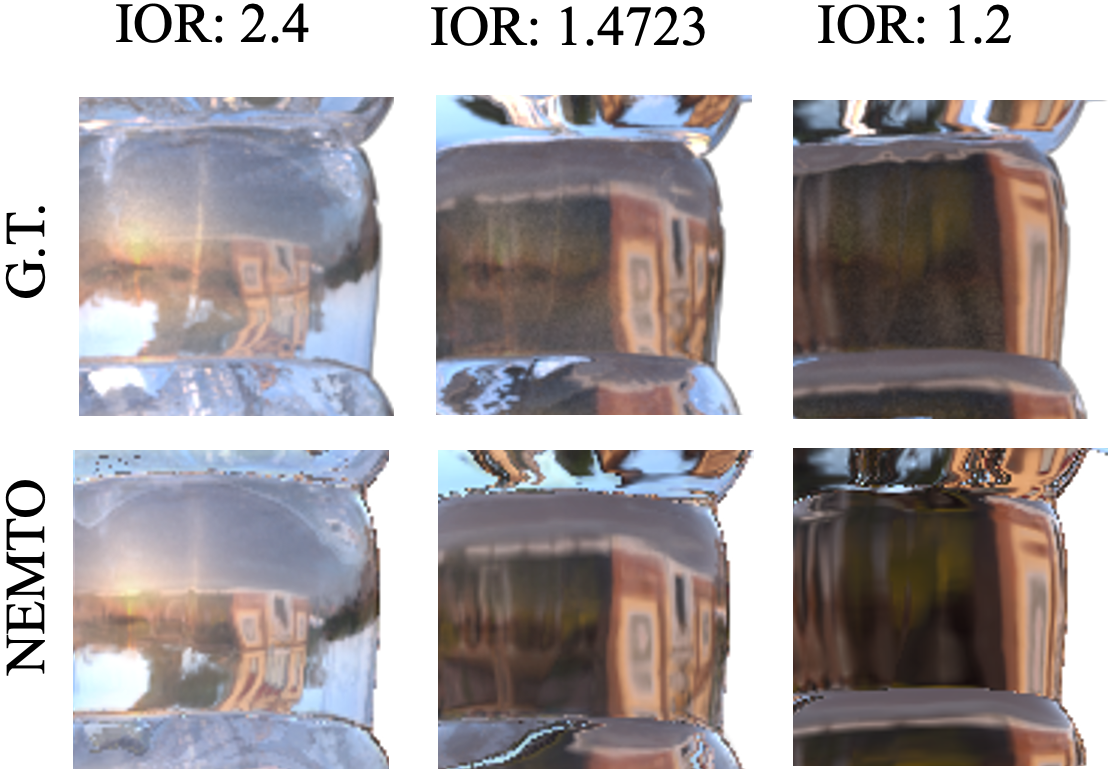}
\end{center}
   \caption{\textbf{Experiments on different transparent media. }We show that NEMTO works for different transparent media other than glass. The learned $\eta_\mathbf{t}$ is adaptive to different media and allows our model to synthesize faithful results.}
\label{fig:ablation-ior}
\vspace{-0.4cm}
\end{figure}

\vspace{-0.2cm}
\medskip
\noindent\textbf{Robustness to different refractive indices.}
We conducted experiments on transparent objects rendered with various IORs to showcase the robustness of our framework to IOR changes. Our approach is suitable for different types of refractive materials, as demonstrated in Fig.~\ref{fig:ablation-ior}. Note that our predicted $\eta_\mathbf{t}$ for the blending of ray refraction and reflection is also adaptive to different IOR, as shown in the case for IOR = 2.4, the reflected radiance is adequately brighter than in IOR = 1.4723 and 1.2.

\begin{table}
\scalebox{0.75}{
  \centering
  \begin{tabular}{@{}lcccccc@{}}
  
    \toprule
     & \multicolumn{3}{c}{ Novel View } 
     & \multicolumn{3}{c}{Relighting} \\
     
    \cmidrule(lr){2-4}\cmidrule(lr){5-7}
    
    Method 
    & PSNR $\uparrow$ & SSIM $\uparrow$ & LPIPS $\downarrow$ 
    & PSNR $\uparrow$ & SSIM $\uparrow$ & LPIPS $\downarrow$ \\
    
    \cmidrule(lr){2-4}\cmidrule(lr){5-7}
    
    NeRF~\cite{mildenhall2020nerf} 
    & 21.274 & 0.837 & 0.171
    & - & - & - 
    \\ 

    Eikonal~\cite{bemana2022eikonal} 
    & 15.866 & 0.452 & 0.589
    & - & - & - 
    \\

    IDR~\cite{yariv2020multiview} 
    & 22.695 & 0.851 & 0.152
    & - & - & - 
    \\ 
    
    PhySG~\cite{zhang2021physg}
    & 19.981 & 0.791 & 0.203
    & 15.412 & 0.749 & 0.237
    \\ 
        
    \midrule
    
    SDF-A 
    & 21.758 & 0.828 & 0.145
    & 17.846 & 0.787 & 0.192
    \\ 

    w/o $\mathcal{L}_{\textrm{rg}}$
    & 15.659 & 0.746 & 0.221
    & 14.585 & 0.713 & 0.238
    \\
    
    w/o $\mathcal{L}_{\textrm{rs}}$
    & 21.623 & 0.811 & 0.163
    & 19.026 & 0.823 & 0.149
    \\
    
    \textbf{NEMTO} 
    & \textbf{26.582} & \textbf{0.924} & \textbf{0.083}
    & \textbf{25.147} & \textbf{0.918} & \textbf{0.098}
    \\
    
    \bottomrule
  \end{tabular}
  }
  \vspace{0.05cm}
  \caption{\textbf{Quantitative Evaluations. } We present the average result on all synthetic datasets. The first three methods are not capable of relighting. Our method performs significantly better on both novel view and relighting synthesis than all of our baseline methods and ablation experiments. }
  \label{tab:quanti}
  \vspace{-0.4cm}
\end{table}

\begin{figure}
\begin{center}
   \includegraphics[width=0.9\linewidth]{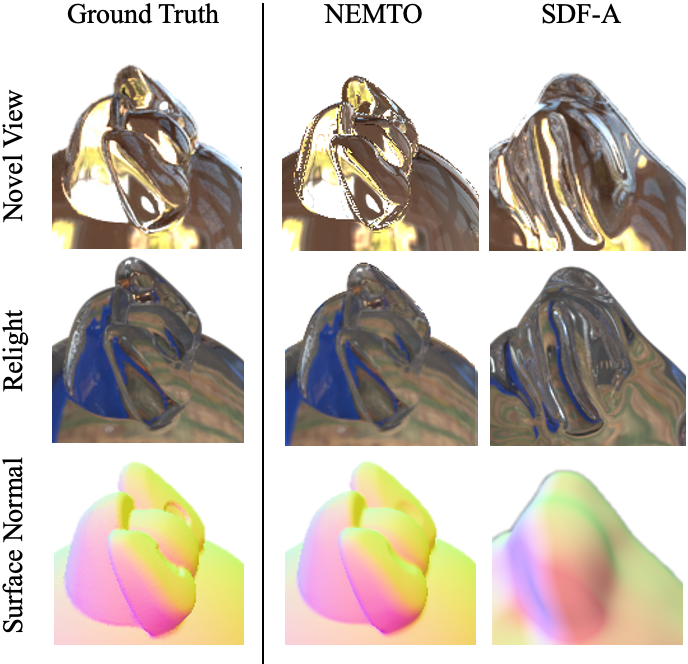}
\end{center}
\vspace{-0.3cm}
   \caption{\textbf{Qualitative ablation on SDF-A}. SDF-A shows that jointly optimizing refraction and geometry is prone to error. Our approach performs significantly better than this naive approach.}
\label{fig:ablation-sdfa}
\vspace{-0.4cm}
\end{figure}

\begin{figure}[t]
\begin{center}
   \includegraphics[width=0.9\linewidth]{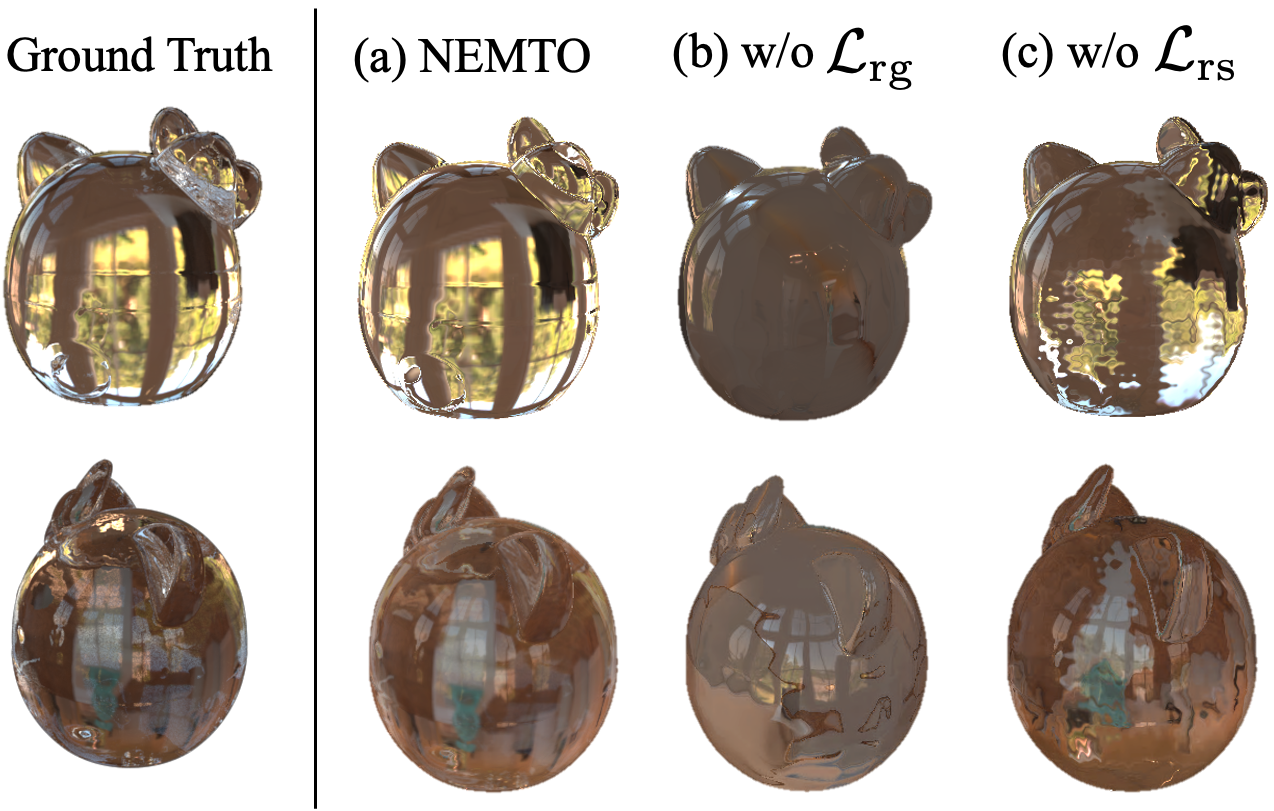}
\end{center}
\vspace{-0.4cm}
   \caption{\textbf{Ablation on losses for ray refraction optimizations.} Each experiment is trained with a frozen geometry network to demonstrate the effect of each loss term on ray bending. }
\label{fig:ablation-loss}
\vspace{-0.4cm}
\end{figure}

\vspace{-0.2cm}
\subsection{Ablation Studies}
We perform the following ablation studies to demonstrate the effectiveness of Our RBN, $\mathcal{L}_{\textrm{rg}}$, and $\mathcal{L}_{\textrm{rs}}$. 

\vspace{-0.2cm}
\medskip
\noindent\textbf{Ablation on ray bending network.} We implemented a naive version of our method~\textbf{SDF-A} without using RBN. It renders transparent objects with \textit{analytical} refraction to demonstrate the effectiveness of our RBN and neural environment matting method over the use of a physically-based differentiable renderer on transparent objects. As shown in Fig.~\ref{fig:ablation-sdfa}, our method synthesizes more accurate results when jointly optimizing for geometry and light refraction, which are better disentangled. This is evident from the smoother surfaces of our method due to $\mathcal{L}_{\textrm{rs}}$. NEMTO estimated smoother surface normal than SDF-A, and gives much more faithful ray refractions.

\begin{figure}[t]
\begin{center}
   \includegraphics[width=\linewidth]{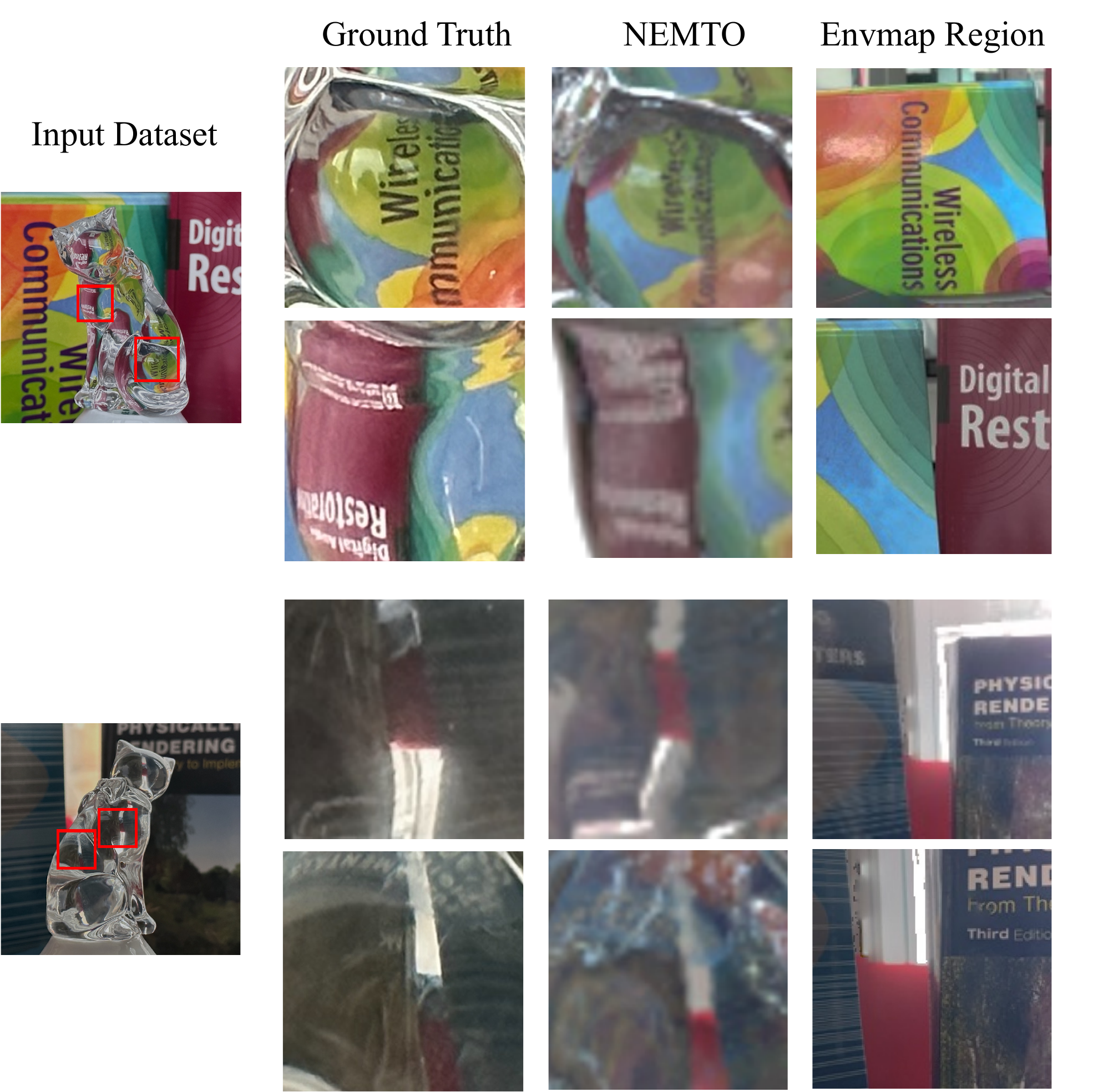}
\end{center}
   \caption{\textbf{Qualitative results on novel view synthesis for real-world captured data.} We show the novel view synthesis of NEMTO on our captured real-world dataset compared to ground truth. The rightmost column displays chosen regions from the environment map as a reference for their corresponding refractions in our synthesis.}
\label{fig:rw-cat}
\end{figure}

\begin{table}[t]
  \centering
    \scalebox{0.9}{
      \begin{tabular}{@{}lcccc@{}}
        \toprule
         Method & PSNR $\uparrow$ & SSIM $\uparrow$ & LPIPS $\downarrow$ \\
        \cmidrule{2-4}
        NStudio & 24.67 & 0.79 &  0.23 \\
        \textbf{NEMTO} & \textbf{27.34} & \textbf{0.85} & \textbf{0.17} \\
        \bottomrule
        \vspace{-0.2cm}
      \end{tabular}
    }
   \caption{\textbf{Quantitative comparison between NeRFStudio~\cite{tancik2023nerfstudio} and NEMTO on novel view synthesis}. We evaluate both methods on our captured cat dataset NEMTO renders better results due to its better capture of refractions. }
  \label{tab:rw-cat}
\end{table}

\medskip
\noindent\textbf{Ablation on $\mathcal{L}_{\textrm{rg}}$ and $\mathcal{L}_{\textrm{rs}}$.} For experiments on the refraction guiding and refraction smoothness loss, we fix the optimized geometry and only show different optimization results for refraction prediction. The lower part of Tab.~\ref{tab:quanti} shows quantitative evaluation that our complete architecture performs better than without each of these loss terms. Fig.~\ref{fig:ablation-loss} compares the learned refraction from each ablation experiment: in column (b) without $\mathcal{L}_{\textrm{rg}}$, the model cannot learn the correct direction; in column (c) without $\mathcal{L}_{\textrm{rs}}$, the optimized ray refraction is around the true scope but shows wrong wave-patterned artifacts.

\subsection{Real World Data Evaluation}
\medskip
\noindent\textbf{Datasets.} For real-world evaluation, we have two sets of experiments. Firstly, we use 4 datasets of transparent objects with complex geometry (dog, monkey, pig, and mouse shape) from TLG ~\cite{li2020through}. As TLG only provides 10-12 images for each object, we render training data using the provided ground truth CT-scanned meshes and environment illuminations. Our evaluations are performed on their released real-world images. Details of all our dataset generation procedures can be found in the supplementary. 

Additionally, we captured a real-world transparent object with sufficient training images to assess the applicability of NEMTO on real-world captured data. Details of our dataset capture procedure are in the supplementary materials.

\begin{figure}[t]
\begin{center}
\includegraphics[width=\linewidth]{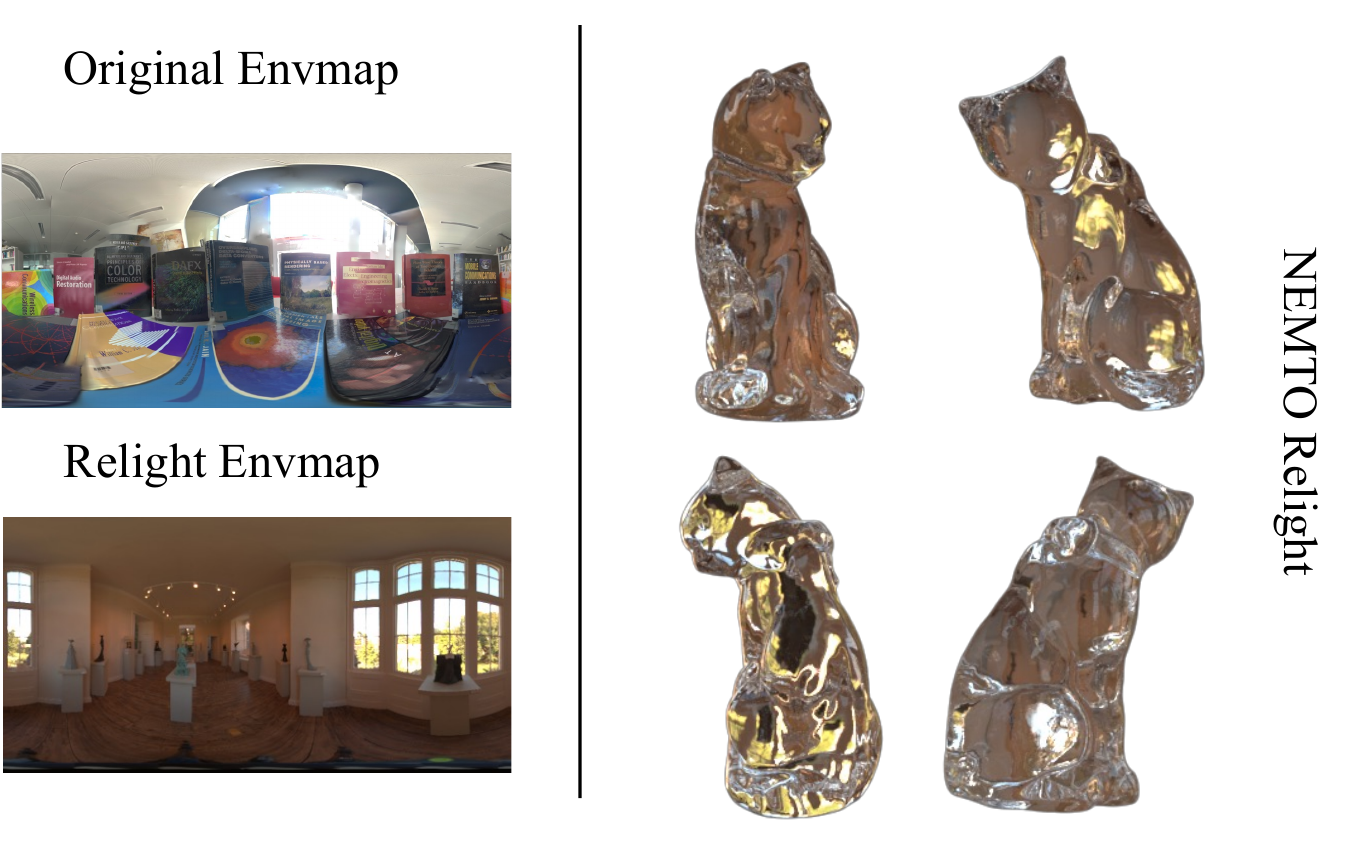}
\end{center}
   \caption{\textbf{Qualitative results on relighting synthesis for real-world captured data}. Our captured environment map is shown in the top left. NEMTO can render visually-plausible relighted transparent objects. }
\label{fig:rw-rl}
\end{figure}

\medskip
\noindent\textbf{Qualitative results on the captured dataset. } Fig.~\ref{fig:rw-cat} and~\ref{fig:rw-rl} show the rendered results from NEMTO trained on captured real-world data. Despite the inaccuracy in real-world camera poses and captured environment maps, NEMTO synthesizes faithful and visually-plausible novel views and relighting results. In Fig.~\ref{fig:rw-cat}, we select geometric regions that produce approximately two bounces of refraction and compare synthesized refractions with the corresponding sections on the Envmap. From these comparisons, we show that our refraction synthesis is quite realistic. 

\begin{figure*}
\begin{center}
\includegraphics[width=.95\linewidth]{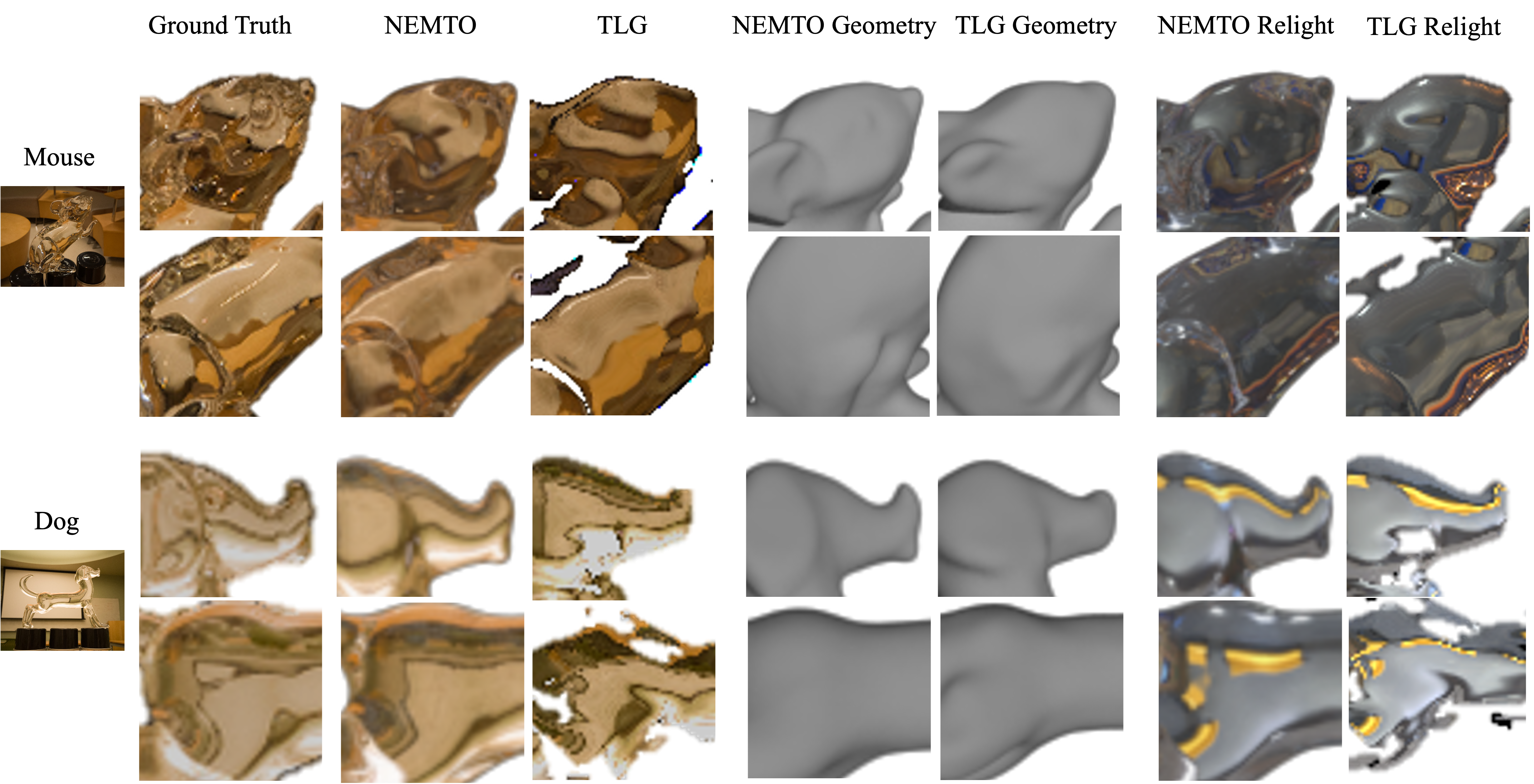}
\end{center}
   \caption{\textbf{Qualitative results on image synthesis and extracted geometry for `rendered' real-world data. } We compare our extracted geometry, novel view synthesis, and relighting with the extracted geometry and rendering layer of TLG~\cite{li2020through}, which restricts the light bounce within transparent media to only two bounces. }
\label{fig:relightrw}
\end{figure*}

\medskip
\noindent\textbf{Quantitative results on the captured dataset. } To quantitatively evaluate the performance of NEMTO on the captured real-world dataset, we provide a comparison on the quality of novel view synthesis between NEMTO and the combined implementation `nerfacto' from several SOTA NeRF models~\cite{barron2021mip, barron2022mipnerf360, martin2021nerf, muller2022instant} from NeRFStudio~\cite{tancik2023nerfstudio} in Tab.~\ref{tab:rw-cat}. NEMTO renders high-frequency refraction details that are closer to ground truth than `nerfacto' which cannot work with light transmission.

\medskip
\noindent\textbf{Qualitative results on the `rendered' real-world datasets. } 
Fig.~\ref{fig:relightrw} shows our results for real-world images trained with synthetic data. Our method is able to predict accurate ray refraction for transparent objects and produces a smoother surface normal prediction on geometry extraction than TLG. TLG designed a novel differentiable rendering layer for physically-based transparent object modeling, but it only renders up to two bounces of refraction, whereas our method does not pose an upper bound on the number of ray bounces. Moreover, TLG does not work with an unknown IOR for transparent objects. Note that, although TLG claims to require only 10-12 images for testing, it requires rendering a large-scale synthetic dataset with 1.5k HDR (High Dynamic Range) envmaps for training, which is unnecessary in our case.

\section{Limitations and Conclusion}
\label{sec:conc}
\medskip
\noindent\textbf{Limitations.} There are a few limitations to our work. First, although NEMTO does not assume homogeneous IOR of transparent media, there is no explicit supervision for heterogeneous transparent objects. With the introduction of appropriate loss functions such as from Eikonal Rendering~\cite{ihrke2007eikonal}, we believe NEMTO can be extended for a wider variety of transparent media. Secondly, our model requires a preprocessing of image data for environment illumination and object masks, as we cannot jointly optimize illumination along with geometry and object appearance for transparent objects. Lastly, NEMTO focuses on unpolarized transparent objects and does not provide experiments on polarized transparent media~\cite{cui2017polarimetric, 5539828, 1467539}.  

\medskip
\noindent\textbf{Conclusion.} We have presented NEMTO, an end-to-end pipeline for novel view and relighting synthesis of transparent objects with complex geometry. Our method jointly optimizes geometry and highly illumination-dependent object appearance and generates high-quality synthesis. 

\section{Acknowledgement}
This work was supported in part by the Swiss National Science Foundation via the Sinergia grant CRSII5-180359. The authors thank Ziyi Zhang for his technical support at the early stage of this work, and thank Yufan Ren, Ehsan Pajouheshgar, Martin Everaert, Bahar Aydemir, Deblina Bhattacharjee, Michele Vidulis, and Merlin Nimier-David for their time spent on proof-reading and kind suggestions during the paper writing.


{\small
\bibliographystyle{ieee_fullname}
\bibliography{egbib}
}

\end{document}